\documentclass[conference]{IEEEtran}
\IEEEoverridecommandlockouts
\usepackage{cite}
\usepackage{amsmath,amssymb,amsfonts}
\usepackage{algorithmic}
\usepackage{graphicx}
\usepackage{textcomp}
\usepackage{xcolor}
\usepackage{hyperref}
\usepackage{tcolorbox}
\usepackage{listings}

\lstdefinestyle{simplecode}{
    basicstyle=\ttfamily\small,
    breaklines=true,
    showspaces=false,
    showstringspaces=false,
    showtabs=false,
    tabsize=2,
    frame=single,
    frameround=tttt,
    backgroundcolor=\color{gray!10},
    rulecolor=\color{orange!40!black}
}

\newenvironment{toolresponse}
{\par\noindent\begin{tcolorbox}[colback=orange!5, colframe=orange!40, title=Tool Response, fontupper=\small, before skip=3pt, after skip=3pt]
\begin{center}\begin{minipage}{0.95\textwidth}\ttfamily\small\obeylines\obeyspaces}
{\end{minipage}\end{center}\end{tcolorbox}\par\noindent}

\newenvironment{userquery}
{\par\noindent\begin{tcolorbox}[colback=gray!10, colframe=gray!40, title=User Query, fontupper=\small, before skip=3pt, after skip=3pt]}
{\end{tcolorbox}\par\noindent}

\newenvironment{systemresponse}
{\par\noindent\begin{tcolorbox}[colback=blue!5, colframe=blue!30, title=Agent Response, fontupper=\small, before skip=3pt, after skip=3pt]}
{\end{tcolorbox}\par\noindent}

\def\BibTeX{{\rm B\kern-.05em{\sc i\kern-.025em b}\kern-.08em
    T\kern-.1667em\lower.7ex\hbox{E}\kern-.125emX}}

\begin{document}

\title{Towards Conversational AI for Human-Machine Collaborative MLOps}

\makeatletter
\newcommand{\linebreakand}{%
  \end{@IEEEauthorhalign}
  \hfill\mbox{}\par
  \mbox{}\hfill\begin{@IEEEauthorhalign}
}
\makeatother

\author{
\IEEEauthorblockN{1\textsuperscript{st} George Fatouros}
\IEEEauthorblockA{
\textit{Innov-Acts Ltd}\\
Nicosia, Cyprus}
\IEEEauthorblockA{\textit{Dept. of Digital Systems} \\
\textit{University of Piraeus}\\
Piraeus, Greece}
\and
\IEEEauthorblockN{2\textsuperscript{nd} Georgios Makridis}
\IEEEauthorblockA{\textit{Dept. of Digital Systems} \\
\textit{University of Piraeus}\\
Piraeus, Greece}
\and
\IEEEauthorblockN{3\textsuperscript{rd} George Kousiouris}
\IEEEauthorblockA{\textit{Dept. of Informatics and Telematics} \\
\textit{Harokopio University}\\
Athens, Greece}
\linebreakand
\IEEEauthorblockN{4\textsuperscript{th} John Soldatos}
\IEEEauthorblockA{
\textit{Innov-Acts Ltd}\\
Nicosia, Cyprus}
\and
\IEEEauthorblockN{5\textsuperscript{th} Anargyros Tsadimas}
\IEEEauthorblockA{\textit{Dept. of Informatics and Telematics} \\
\textit{Harokopio University}\\
Athens, Greece}
\and
\IEEEauthorblockN{6\textsuperscript{th} Dimosthenis Kyriazis}
\IEEEauthorblockA{\textit{Dept. of Digital Systems} \\
\textit{University of Piraeus}\\
Piraeus, Greece}
}

\maketitle

\begin{abstract}
This paper presents a Large Language Model (LLM) based conversational agent system designed to enhance human-machine collaboration in Machine Learning Operations (MLOps). We introduce the Swarm Agent, an extensible architecture that integrates specialized agents to create and manage ML workflows through natural language interactions. The system leverages a hierarchical, modular design incorporating a KubeFlow Pipelines (KFP) Agent for ML pipeline orchestration, a MinIO Agent for data management, and a Retrieval-Augmented Generation (RAG) Agent for domain-specific knowledge integration. Through iterative reasoning loops and context-aware processing, the system enables users with varying technical backgrounds to discover, execute, and monitor ML pipelines; manage datasets and artifacts; and access relevant documentation, all via intuitive conversational interfaces. Our approach addresses the accessibility gap in complex MLOps platforms like Kubeflow, making advanced ML tools broadly accessible while maintaining the flexibility to extend to other platforms. The paper describes the architecture, implementation details, and demonstrates how this conversational MLOps assistant reduces complexity and lowers barriers to entry for users across diverse technical skill levels.
\end{abstract}


\section{Introduction}
\label{sec:1}

In today’s rapidly evolving landscape of artificial intelligence (AI) and machine learning (ML), the synergy between human expertise and AI capabilities is increasingly crucial \cite{kiourtis2024xr5}. By combining human understanding and creativity with AI’s computational power and efficiency, organizations can drive more effective decision-making and innovative solutions across a range of domains, including healthcare, finance, and scientific research \cite{fatouros2024can, makridis2020predictive, fatouros_deepvar_2023}.

As ML workflows become more complex, MLOps platforms play a critical role in streamlining the development, deployment, and management of these workflows \cite{zarour2025mlops}. Kubeflow\footnote{https://www.kubeflow.org/} is a leading open-source solution for orchestrating ML pipelines on Kubernetes \cite{george2022end}. However, its feature set can introduce complexities that challenge data scientists and users without extensive  experience on systems and devops processes. This highlights the broader need to make advanced ML platforms more accessible, ensuring that diverse user groups can take advantage of their functionalities.

To address these challenges, we present the \emph{Swarm Agent}, a Large Language Model (LLM) based conversational assistant with access to tools and services to manage ML workflows and data operations required in the ML life cycle. Our implementation integrates with Kubeflow using tools such as the Kubeflow Pipelines (KFP) client, MinIO client, and a vector database, enabling ML pipeline creation and execution, data operations, and context-aware guidance from natural language querying. The design of our assistant is extensible by integrating additional tools or agents and can be adapted to other MLOps platforms, offering a flexible approach to simplifying and expanding access to  advanced ML operations beyond a single technology stack.

This paper presents the design and implementation of our LLM-powered chat assistant, highlighting its role in enhancing human–AI collaboration in ML pipeline management. We describe the system’s architecture and illustrate how each component contributes to a more intuitive user experience. We also present indicative outcomes of the presented system which demonstrate how we reduce complexity for diverse technical skill levels, ultimately lowering barriers to entry for advanced ML tools and driving broader innovation in the field.

The structure of this paper is organized as follows: Section~\ref{sec:2} reviews related work on Conversational Agents (CAs), MLOps platforms and applications of LLMs agents. Section~\ref{sec:system_architecture} presents the architecture of the proposed agentic system along followed by Section~\ref{sec:implementation_details} discussing the implementation details of the system. Section~\ref{sec:evaluation} provides real-world use cases and indicative responses from the Swarm Agent. Section~\ref{sec:conclusion} concludes the paper with key insights and our plans for future work.

\section{Related Works}
\label{sec:2}

\subsection{Conversational Agents and Natural Language Interfaces}

The field of CAs has experienced rapid evolution, transitioning from early rule-based systems to contemporary LLM-driven applications such as ChatGPT and Perplexity. These advancements facilitate more natural and context-aware interactions, significantly broadening the scope of chatbot applications to include complex technical systems \cite{schobel2024charting}. Despite the extensive internal knowledge embedded within LLMs, advanced agent-based applications leverage external tools, including online search engines, code interpreters, and function-calling mechanisms \cite{ouaddi2024architecture} to enhance their utility. Function calling, in particular, enables programmatic retrieval of data from external services via Application Programming Interfaces (APIs), allowing on-demand access to information relevant to user queries and triggering corresponding actions \cite{OpenAIFunctionCalling}. One prominent example of integrating external knowledge into LLM-based agents is Retrieval-Augmented Generation (RAG) which utilizes external knowledge bases to store information that may not be present in the training data of an LLM \cite{lewis2020retrieval}. This method grounds LLM responses, mitigates hallucinations, and allows accurate resolution of user queries using domain-specific, proprietary, or dynamically updated data \cite{fatouros2025marketsenseai}. These developments form the foundational techniques inspiring our approach to managing ML workflows through natural language interactions.

\subsection{MLOps Platforms and Pipeline Orchestration}

The advent of MLOps has transformed the deployment and management of ML workflows. Recent surveys and technical reports detail the evolution of MLOps frameworks, identifying both their transformative potential and the operational complexities they introduce \cite{Shin2025DriverIS}. Kubeflow has emerged as the leading platform for orchestrating ML pipelines on Kubernetes, enabling end-to-end machine learning workflows \cite{bergmanncomparative}. However, its extensive feature set can pose significant challenges for users without a strong technical background, especially in the area of systems and software management. For example, industry experts who have valuable business or domain knowledge often struggle with the technical complexities of MLOps platforms, while data scientists may lack the DevOps expertise needed to efficiently deploy and monitor models in production environments \cite{Keusch2024PlatformAgnosticMO, katevas2024embedding}. This skill gap is particularly evident when industrial specialists attempt to integrate AI solutions into existing workflows but encounter barriers due to the difficulty of handling such technologies \cite{Grilo2024DataAE}. 

Beyond Kubeflow, other MLOps frameworks like Apache Airflow \cite{harenslak2021data}, Dagster, and MLflow \cite{zaharia2018accelerating} are increasingly used to orchestrate ML workflows. While our approach is initially integrated with Kubeflow, it is modular and designed to extend to any platform offering APIs for managing experiments, runs, and logging. MLflow, for example, offers robust experiment tracking and model management APIs; Dagster supports detailed logging and monitoring; and Airflow provides a REST API for workflow orchestration. By leveraging these interfaces, our system can enable conversational management of ML workflows across diverse MLOps environments.

\subsection{Integration of LLMs in Operational Tools}
The integration of agentic AI applications into coding and DevOps has led to the development of tools that enhance software development, productivity and operational workflows \cite{peng2023impact}. GitHub Copilot \cite{GitHubCopilot}, Cursor AI \cite{CursorAIEditor} and Devin AI \cite{DevinAI}, are prominent LLM-powered coding tools that assist developers by suggesting whole lines or entire functions within their editor, streamlining the coding process. CodeGPT offers AI agents tailored for software development teams, providing features such as AI coding assistance and automated code reviews, thereby enhancing productivity through deep codebase understanding \cite{CodeGPT}. These applications exemplify the transformative impact of agentic AI in coding and DevOps, aligning with our proposed system's goal of integrating conversational AI agents into MLOps platforms to facilitate streamlined workflow management.

To the best of our knowledge, no existing system offers a conversational, agentic interface that seamlessly integrates with MLOps platforms like Kubeflow through modular, tool-specific agents that enable ML workflow management through interactive, LLM-driven planning and execution. This fills a critical gap in making complex MLOps environments accessible and intuitive for users of varying technical backgrounds.

\section{System Architecture}
\label{sec:system_architecture}
The proposed system, Swarm Agent, facilitates agentic interaction between users and complex MLOps environments, with a particular focus on Kubeflow. The architecture leverages a modular, extensible design that empowers users to manage data, pipelines, and domain-specific knowledge through natural language leveraging the context understanding of LLMs. The architecture of the Swarm Agent is depicted in Fig. \ref{fig:arch} with the following subsections analyzing the main architectural components.

\begin{figure*}[htbp]
\centerline{\includegraphics[width=\textwidth]{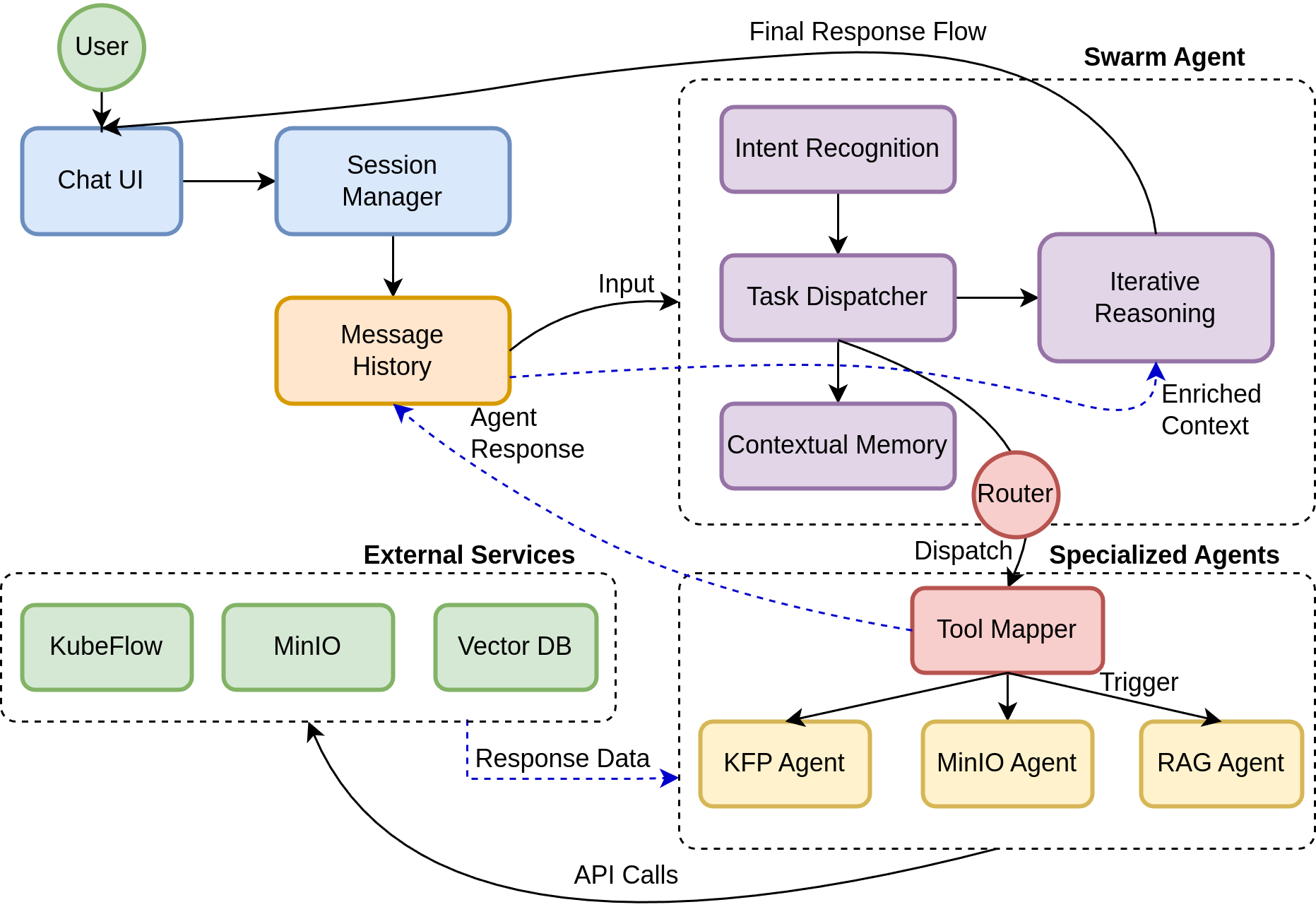}}
\caption{Architecture of the Swarm Agent system for conversational MLOps.}
\label{fig:arch}
\end{figure*}

\subsection{Chat UI and Session Management}

The user interface (Chat UI) serves as the primary human-computer interaction (HCI) endpoint in the system, leveraging asynchronous communication to capture user input and render agent-generated responses. The system implements a session-based architecture to maintain conversational context and user state throughout interactions. The \emph{Session Manager} encapsulates this functionality by maintaining a persistent message history for each user session, tracking thread identifiers, and preserving user authentication context across interactions. This stateful approach enables context-aware reasoning and personalized responses tailored to the user's ongoing conversation flow, creating a continuous and coherent interaction experience.

\subsection{Swarm Agent Core Architecture}
The Swarm Agent serves as a central LLM-powered controller, enabling intelligent coordination across domain-specific agents through a combination of intent recognition, contextual memory, and dynamic orchestration. The architecture comprises three main components: an Intent Recognition System that analyzes user queries to infer the underlying goals and extract relevant parameters; a Contextual Memory Module that maintains a structured representation of the conversation history, enabling coherent multi-turn interactions and reference resolution via a message history structure; and a Task Dispatcher that activates and communicates with downstream agents through a function-calling mechanism.

The system implements hierarchical orchestration by parsing streaming LLM outputs in real-time to detect tool invocation intents. These tool calls are dynamically routed to specialized agents using a tool mapper that matches descriptions from the LLM to their corresponding implementation code. Once agents return their outputs, the Swarm Agent re-incorporates them into the conversation context, enabling iterative refinement of the response.

This architecture supports iterative reasoning loops, where the agent responses are not immediately surfaced to the user but are first contextualized as function call results within the conversation history. The enriched context is used to prompt the LLM again, allowing for recursive reasoning and synthesis across multiple cycles. This mechanism enhances the system’s ability to decompose complex queries into subtasks, handle failures gracefully (e.g., by retrying with adjusted parameters or switching tools), and integrate intermediate outputs into a coherent and high-quality final response.

By combining real-time responsiveness with contextual continuity and recursive processing, the Swarm Agent enables intelligent behavior that goes beyond the capabilities of any individual component. This is particularly impactful in MLOps and technical support contexts, where it must fluidly combine code, system-level reasoning, and natural language explanation.

\subsection{KFP Agent}
The KFP Agent (Fig.~\ref{fig:kfp}) serves as the primary interface between the conversational system and Kubeflow enabling its operation with natural language. This agent provides a complete set of functionalities that encapsulate the lifecycle of ML pipeline management through specialized functions.

\begin{figure}[htbp]
\centerline{\includegraphics[width=0.5\textwidth]{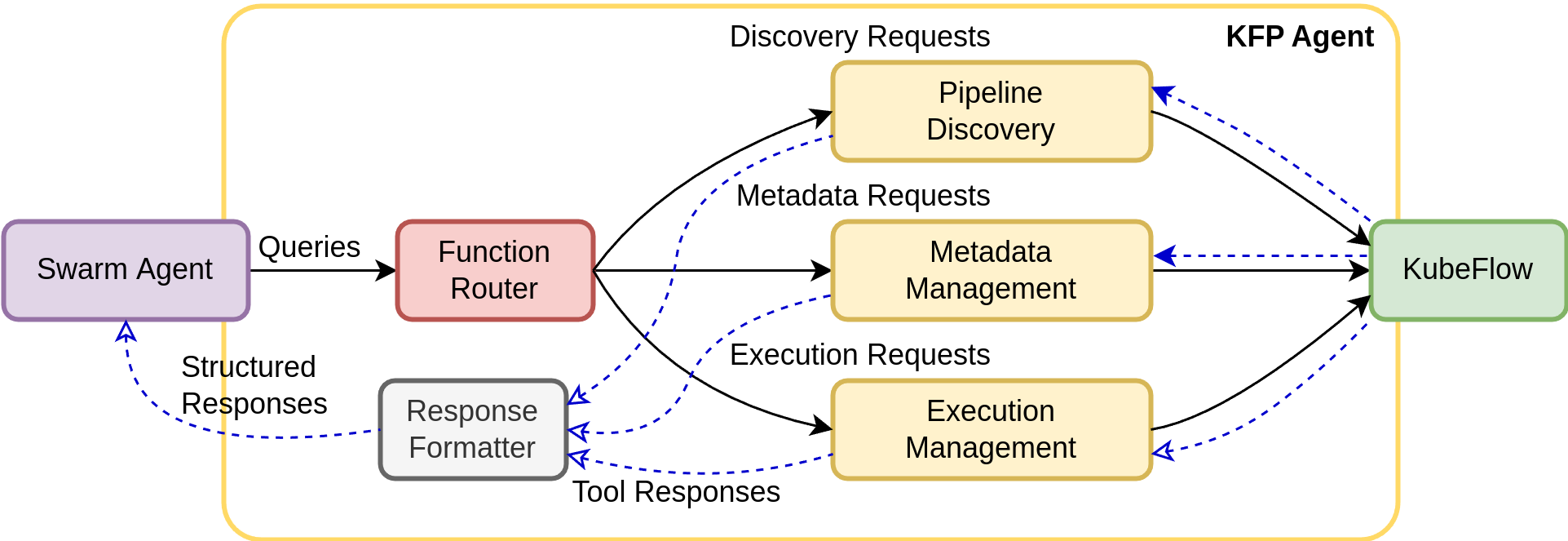}}
\caption{Internal Architecture of the KubeFlow Pipelines (KFP) Agent.}
\label{fig:kfp}
\end{figure}

\subsubsection{Pipeline Discovery and Metadata Management}
The agent implements search capabilities through the \texttt{get\_pipelines} function, allowing users to discover available pipelines using flexible search terms, namespace filtering, and pagination.  For detailed inspection, the agent provides access to pipeline metadata through the \texttt{get\_pipeline\_details} function, which retrieves comprehensive information including descriptions, creation timestamps, versioning information, and technical specifications. The \texttt{get\_pipeline\_version\_details} function additionally extracts pipeline components and parameters from the pipeline specification, offering deeper insights into workflow structure. This detailed metadata extraction includes component definitions, input parameters with their default values and types, enabling users to understand the capabilities and requirements of a pipeline before execution.

\subsubsection{Execution Management and Monitoring}
The KFP Agent facilitates pipeline execution through dedicated functions including \texttt{create\_experiment} for setting up experiment contexts and \texttt{run\_pipeline} for initiating workflow execution. The execution function accepts multiple parameters including experiment IDs, job names, parameter mappings, and pipeline identifiers. The system also provides the \texttt{get\_pipeline\_id} utility to retrieve pipeline identifiers from names, simplifying the execution process for users.
For monitoring purposes, the agent provides visibility into pipeline executions through functions like \texttt{list\_runs} and \texttt{get\_run\_details}, supporting filtered queries by status, experiment context, or search terms. The implementation includes capabilities for real-time status tracking through state information providing information about run status and associated pipeline metadata, providing users with comprehensive visibility into their ML workflows through the Chat UI.

\subsection{MinIO Agent}
The MinIO Agent serves as the data management counterpart to the KFP Agent, facilitating interaction with object storage services that house datasets, models, and artifacts throughout the ML lifecycle as depicted in Fig.~\ref{fig:minio}. This agent enables users to explore, analyze, and leverage stored data assets including metrics and results from the ML experiments through text queries.

\begin{figure}[htbp]
\centerline{\includegraphics[width=0.5\textwidth]{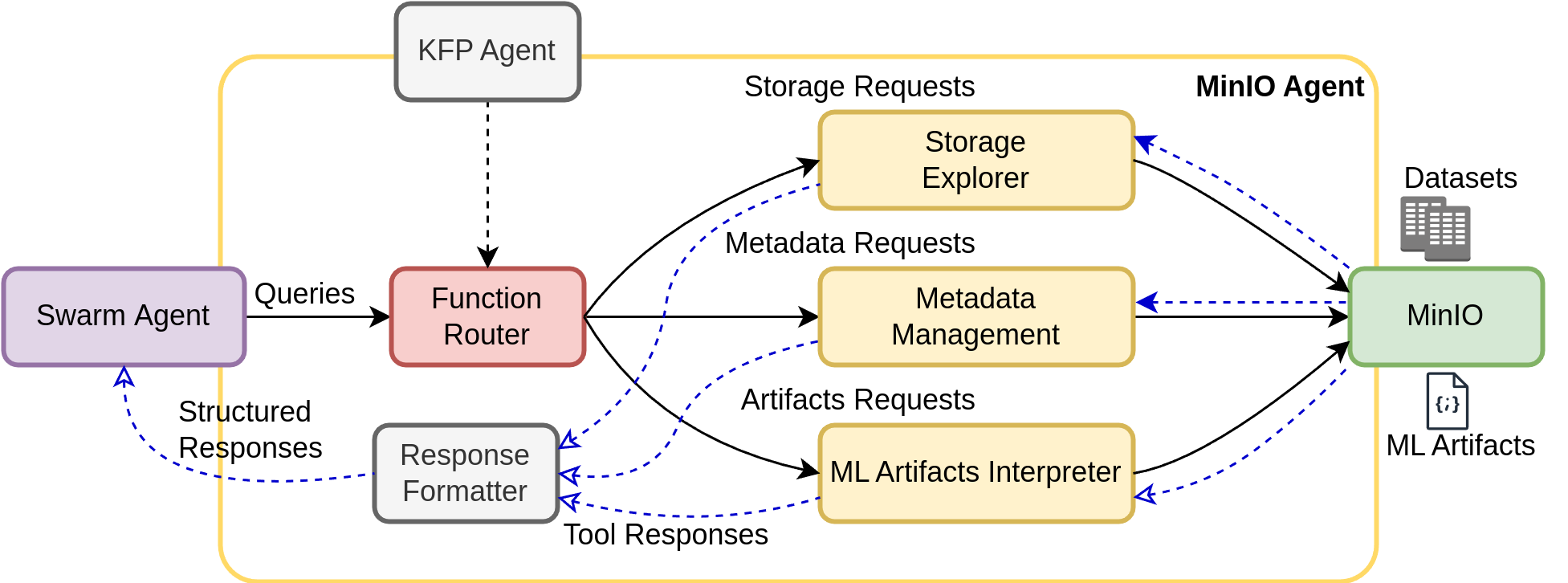}}
\caption{Internal Architecture of the MinIO Agent.}
\label{fig:minio}
\end{figure}

\subsubsection{Storage Exploration and Discovery}
The MinIO Agent implements bucket and object discovery through the \texttt{get\_minio\_info} and \texttt{list\_user\_buckets} functions, allowing users to navigate through storage hierarchies using text queries. These function accept parameters like bucket name and prefix to enable flexible exploration of storage contents. Key to this approach is the LLM's ability to interpret abstract, intent-based requests and translate them into precise parameters for the MinIO client. For example, when a user asks "Show me recent image datasets from last month's experiments," the LLM maps this request to appropriate bucket, prefix, and filter parameters without requiring the user to understand storage structures or query syntax. 

\subsubsection{ML Artifacts Interpretation}
A key capability of the MinIO Agent is its ability to retrieve and interpret artifacts generated during ML workflow execution through specialized functions like \texttt{get\_pipeline\_artifacts}, \texttt{get\_model\_metrics}, and \texttt{get\_pipeline\_visualization}. These functions enable access to various artifact types including:

\begin{itemize}
    \item Performance metrics stored as JSON files
    \item Evaluation visualizations such as ROC curves, confusion matrices, and learning curves
    \item Model interpretation plots like feature importance charts and partial dependence plots
    \item Validation results across different data splits or cross-validation folds
\end{itemize}

The agent not only retrieves these artifacts but can also interprete them within the conversation, translating ML training result into actionable insights. This helps users understand model performance patterns, identify potential issues, and formulate improvement strategies. By contextualizing performance metrics within the broader ML workflow, the agent provides guidance on hyperparameter tuning, feature selection refinements, or architectural changes that could improve model effectiveness.

\subsubsection{Integration with ML Workflows}
The MinIO Agent maintains awareness of workflow context through integration with the KFP Agent, enabling it to relate stored artifacts to specific pipeline executions. This contextual awareness allows the agent to:

\begin{itemize}
    \item Connect model outputs to their source pipelines and input datasets
    \item Track version evolution across multiple pipeline runs
    \item Compare performance metrics between workflow iterations
    \item Provide recommendations that consider both execution history and data characteristics
\end{itemize}

By maintaining this workflow context, the MinIO Agent can deliver significantly more valuable insights than would be possible through isolated storage access, effectively serving as the memory layer for the conversational MLOps ecosystem.

\subsection{RAG Agent} 

The RAG Agent implements Retrieval-Augmented Generation to enhance the system's responses with domain-specific knowledge from documentation about reusable Kubeflow workflow components. As depicted in Fig.~\ref{fig:rag}, this agent facilitates context-aware responses by integrating a vector database with the conversational flow.

\begin{figure}[htbp]
\centerline{\includegraphics[width=0.5\textwidth]{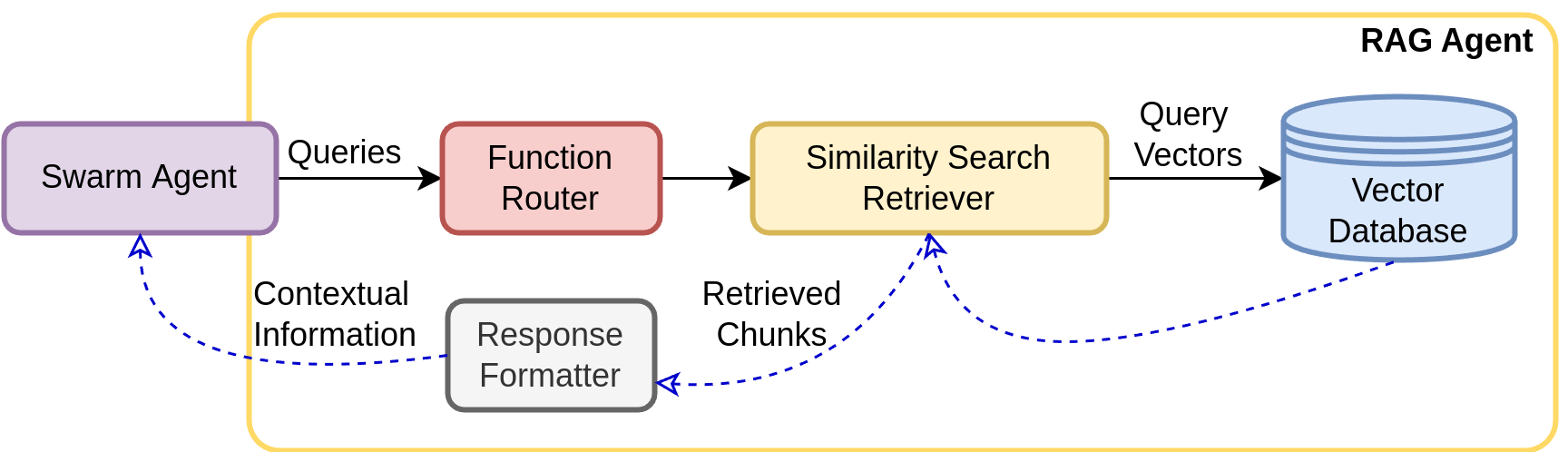}}
\caption{Internal Architecture of the RAG Agent.}
\label{fig:rag}
\end{figure}

\subsubsection{Knowledge Indexing and Embedding}
The RAG Agent employs a data injection pipeline (Fig.~\ref{fig:indexing}) that processes documentation into a searchable knowledge base. This process involves several stages as implemented in the data injection module. First, the system loads the input documents, primarily PDF files containing technical documentation. These documents undergo semantic chunking, which divides texts into coherent segments based on semantic boundaries rather than arbitrary character counts \cite{tomkou2025bridging}. This approach preserves the contextual integrity of information while creating manageable units for retrieval.
Each chunk is then embedded using an embedding model, transforming textual data into high-dimensional vector representations that capture semantic relationships. These vectors are stored in a vector database. The implementation includes metadata preservation, ensuring that each vector maintains associations with its source document title and content for efficient retrieval context.

\begin{figure}[htbp]
\centerline{\includegraphics[width=0.5\textwidth]{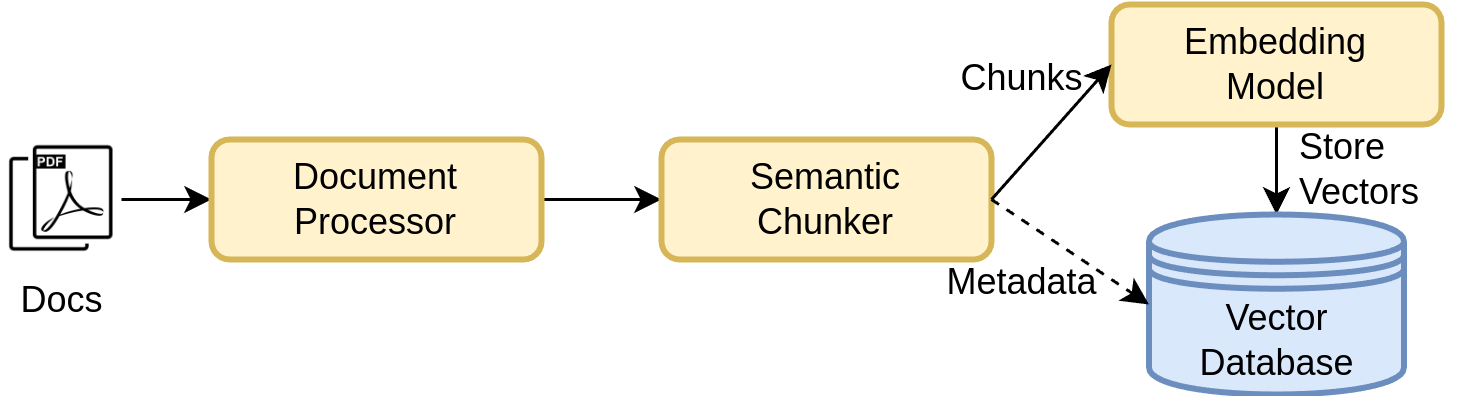}}
\caption{Architecture of the Knowledge Indexing Pipeline.}
\label{fig:indexing}
\end{figure}

\subsubsection{Knowledge Retrieval and Integration}
The core retrieval functionality is implemented through similarity searches against the vector database. This implementation retrieves the most semantically similar document chunks for each query. Retrieved information from the RAG Agent is appended in the message history and integrated into the reasoning loops of the Swarm Agent. This integration enables information synthesis where domain knowledge from documentation is combined with pipeline execution capabilities and data management functions.
The RAG approach is particularly valuable when dealing with proprietary code components or custom Kubeflow extensions that wouldn't be represented in the LLM's pre-trained knowledge. For example, when users inquire about organization-specific pipeline templates, customized workflow components, or proprietary model implementations, the RAG Agent can provide accurate documentation that isn't available in public domains. Thus, this agent ensures that users receive contextually appropriate and factual guidance on Kubeflow components, enhancing the overall collaborative MLOps experience.

\subsection{Error Handling and Response Formatting}
The system incorporates error handling to manage API failures, invalid input or system problems. Each function implements try-catch mechanisms that capture and log detailed error information allowing the LLM to adapt the input parameters to the Agent's functions or request additional information from the user. Responses are structured as serializable dictionaries, facilitating smooth integration with the parent Swarm Agent and ensuring readable information presentation to end users. This consistent response format enables the LLM to reason effectively about operation results and formulate appropriate follow-up responses or actions.

\subsection{Extensibility}

The architecture is inherently modular, enabling the integration of additional specialized agents and external services with minimal changes to the core system. This extensibility can support the incorporation of AI model benchmarking tools to evaluate and compare model performance across different configurations; data quality assessment services that identify issues in training datasets; and specialized code assistants that can generate or modify Kubeflow pipeline definitions using domain-specific languages (DSL) \cite{fragiadakis2024evaluating, fatouros2023comprehensive}. Each new capability is exposed through the function-calling mechanism, maintaining a consistent interaction pattern while expanding system functionality. For instance, a code assistant agent could leverage the KFP client's DSL to generate complete pipeline specifications from user's instructions and then the KFP Agent upload and run it at KubeFlow. This flexibility allows the system to evolve as new capabilities can be integrated without disrupting existing functionality.

\section{Implementation Details}
\label{sec:implementation_details}

The conversational interface is built on Chainlit\footnote{https://github.com/Chainlit/chainlit}, providing an asynchronous communication framework with step-based processing for improved traceability and debugging. For LLM interactions, the system leverages OpenAI's API with structured function calling capabilities. However, any LLM with function calling capabilities can be supported. The RAG component incorporates LangChain's SemanticChunker for context-aware document segmentation \cite{zhao2024meta}, OpenAI's text-embedding-3-small model for generating vector representations, and Pinecone as the vector database. The knowledge retrieval pipeline uses LlamaIndex's VectorIndexRetriever with a top-5 similarity threshold \cite{malviya2024scalability}. For MLOps integrations, the system utilizes the official Kubeflow Pipelines SDK\footnote{https://kubeflow-pipelines.readthedocs.io/} for workflow management and the MinIO Python client for object storage interactions. 
The system implements authentication using Keycloak\footnote{https://www.keycloak.org/} as the Single Sign-On (SSO) solution. During login to the chat interface, the system retrieves user-specific session variables including Kubeflow namespace configurations, MinIO bucket permissions, and KFP client credentials. This approach ensures that users only access resources they are authorized to use while maintaining credential isolation.

\section{Use Cases and Indicative Responses}
\label{sec:evaluation}

To demonstrate the practical utility of the Swarm Agent system in enabling conversational MLOps, we present several representative use cases based on the diabetes classification pipelines implemented in our experimental environment. These examples illustrate how the system facilitates human-machine collaboration across different user skill levels and task complexities.

\subsection{Pipeline Discovery and Exploration}

\begin{userquery}
What ML pipelines are available?
\end{userquery}

When the user inquires about available ML pipelines, the system first leverages the KFP Agent to query the Kubeflow pipeline registry (as shown in the tool response containing structured JSON data). This raw information is then processed through the Swarm Agent, resulting in a well-formatted, human-readable response that organizes the pipeline information hierarchically.\\

\begin{toolresponse}
{
  "content": {
    "total\_pipelines": 2,
    "total\_available": 4,
    "namespace\_type": "shared",
    "namespace": "shared",
    "pipelines": [
      {
        "id": "d74...47559c",
        "name": "diabetes-svm...",
        "description": "A pipeline for...",
        "created\_at": "..."},
      {
        "id": "297...92a78",
        "name": "diabetes-dt...",
        "description": "A pipeline for...",
        "created\_at": "...."}]
  }
}
\end{toolresponse}

\begin{systemresponse}
There are currently 2 ML pipelines available in the system:
\begin{enumerate}
    \item diabetes-svm-classification-pipeline
    \begin{itemize}
        \item Description: A pipeline for diabetes...
        \item Created at: 2025-04-14
    \end{itemize}
    \item diabetes-dt-classification-pipeline
    \begin{itemize}
        \item Description: A pipeline for diabetes...
        \item Created at: 2025-04-14
    \end{itemize}
\end{enumerate}

If you want more details about any specific pipeline or want to explore their runs or results, please let me know!
\end{systemresponse}

\subsection{Pipeline Details and Parameter Understanding}

\begin{userquery}
Tell me more about the diabetes SVM classification pipeline and what parameters it accepts.
\end{userquery}

This example demonstrates the system's iterative reasoning capability and how it uses multiple agent calls to formulate helpful responses. Behind the scenes, the Swarm Agent first uses the KFP Agent to retrieve basic pipeline metadata through the \texttt{get\_pipeline\_details} function, identifying the pipeline ID and available versions. Upon recognizing that more information is needed for a complete response, it makes a second call to retrieve detailed component specifications and parameter definitions via \texttt{get\_pipeline\_version\_details}.\\

\begin{toolresponse}
{
  "id": "d74d8...559c",
  "name": "diabetes-svm-...",
  "description": "A pipeline...",
  "created\_at": "2025-04-14...",
  "pipeline\_versions": [...]
}
\end{toolresponse}

\begin{toolresponse}
{
  "pipeline\_spec": "...",
  "components": {
    "comp-train-svm": {
      "inputDefinitions": {
        "parameters": {
          "C": {"defaultValue": 1, "parameterType": "NUMBER\_DOUBLE"},
          "kernel": {"defaultValue": "rbf", "parameterType": "STRING"},
          "random\_state": {"defaultValue": 42, "parameterType": "NUMBER\_INTEGER"}}}},
    "comp-split-data": {
      "inputDefinitions": {
        "parameters": {
          "random\_state": {"defaultValue": 42, "parameterType": "NUMBER\_INTEGER"},
          "test\_size": {"defaultValue": 0.3, "parameterType": "NUMBER\_DOUBLE"}
        }
      }
    }
  }
}
\end{toolresponse}

\begin{systemresponse}
The Diabetes SVM Classification Pipeline is a demonstration pipeline for diabetes classification using a Support Vector Machine (SVM) model with artifact tracking in MinIO.

Pipeline Structure:
\begin{enumerate}
    \item  Data loading - Loads the diabetes dataset from scikit-learn and converts it to a binary classification problem
    \item Data splitting - Divides data into training and test sets
    \item Preprocessing - Normalizes features using StandardScaler
    \item Model training - Trains an SVM classifier on the preprocessed training data
    \item Evaluation - Generates performance metrics and visualizations
    \item Artifact storage - Saves models, metrics, and visualizations to MinIO
\end{enumerate}

The pipeline accepts the following parameters:
\begin{itemize}
    \item test\_size: float (default: 0.3) - Proportion of the dataset used for testing (e.g., 0.3 means 30\% for testing, 70\% for training)
    \item random\_state: int (default: 42) - Random seed for reproducibility of data splitting and model training
    \item svm\_C: float (default: 1.0) - Regularization parameter; smaller values enforce stronger regularization
    \item svm\_kernel: string (default: 'rbf') - The kernel type to be used in the algorithm, options include 'linear', 'poly', 'rbf', 'sigmoid'
    \item Various MinIO/authentication parameters (typically set by system administrators)
\end{itemize}

Would you like to run this pipeline with specific parameters or compare it with the Decision Tree pipeline?
\end{systemresponse}

\subsection{Comparing Model Performance}

\begin{userquery}
Compare the performance of SVM and Decision Tree models for diabetes classification.
\end{userquery}

This example demonstrates the system's ability to organize parallel tool calls for efficient data retrieval and comparison. Upon receiving the user's query, the Swarm Agent first calls MinIO's \texttt{list\_user\_buckets} function, which returns information about available storage buckets, identifying the paths in MinIO containing the SVM and Decision Tree pipeline artifacts. The system recognizes the two relevant pipeline names: \texttt{diabetes-svm-classification} and \texttt{diabetes-dt-classification}. With this context, the Swarm Agent then makes parallel calls to the \texttt{get\_model\_metrics} function—one for each model type—to retrieve their performance metrics. The system then synthesizes these parallel data streams, aligning the metrics for direct comparison and presenting them to the user.\\

\subsection{Additional Supported Query Types}

The Swarm Agent supports diverse query types beyond the demonstrated examples, enabling comprehensive MLOps management through natural language. 
\begin{systemresponse}
I've compared the performance metrics of the SVM and Decision Tree models for diabetes classification based on the most recent runs:

\begin{center}
\begin{tabular}{|l|c|c|}
\hline
\textbf{Metric} & \textbf{SVM} & \textbf{Decision Tree} \\
\hline
Accuracy & 0.752 & 0.706 \\
Precision & 0.739 & 0.699 \\
Recall & 0.773 & 0.718 \\
F1 Score & 0.756 & 0.709 \\
AUC & 0.842 & 0.708 \\
\hline
\end{tabular}
\end{center}

Confusion Matrices:
\begin{itemize}
\item SVM: [[49, 18], [15, 51]]
\item Decision Tree: [[77, 34], [31, 79]]
\end{itemize}

Key insights:
\begin{enumerate}
\item The SVM model outperforms the Decision Tree model across all key metrics including accuracy, precision, recall, F1 score, and AUC.
\item The higher AUC of the SVM (0.842) indicates better overall discrimination capability between classes.
\item The confusion matrix for SVM shows fewer false positives and false negatives compared to the Decision Tree.
\end{enumerate}

Would you like me to provide visualizations or more detailed analysis of these models' results?
\end{systemresponse}
\newline
These include:
\subsubsection{Experiment Creation and Modification Queries} Users can request "Run a new experiment using sigmoid kernel" leveraging the system's understanding of pipeline parameters and chat history.
\subsubsection{Data Management Queries} The MinIO Agent facilitates queries like "List all datasets uploaded in the past month" or "Compare feature importance across multiple experiments," enabling straightforward data exploration.

\subsubsection{Workflow Monitoring and Debugging} Users can ask "Why did the latest diabetes pipeline run fail?" or "Show me logs from the preprocessing step of run-123," with the system aggregating relevant execution details and presenting actionable insights.
\subsubsection{Domain-Specific Knowledge Integration} The RAG Agent handles queries such as "How does the XYZ classifier handle categorical features in diabetes prediction?" by retrieving and contextualizing documentation from proprietary model libraries not covered in the LLM's pre-trained knowledge.
\subsubsection{Cross-Pipeline Analysis} Complex analytical queries like "Which pipeline version achieved the highest precision for minority class samples?" or "Generate a report comparing all model runs from April" are processed through coordinated function calls across multiple specialized agents.

These capabilities collectively transform how research teams interact with MLOps infrastructure, removing technical barriers and enabling domain experts to directly leverage advanced ML workflows in their research.

\section{Conclusion and Future Work}
\label{sec:conclusion}
This paper presented the Swarm Agent, a novel LLM-powered conversational system for the management of MLOps through chat interactions. By integrating specialized agents for KubeFlow Pipelines orchestration, MinIO data management, and domain-specific knowledge retrieval, our approach bridges the gap between complex ML infrastructure and users with diverse technical backgrounds. Through iterative reasoning and contextual processing, the system enables pipeline discovery, execution, monitoring, and results analysis without requiring expertise in underlying technologies. The demonstrated examples illustrate how this conversational approach reduces cognitive load, eliminates technical barriers, and enhances human-machine collaboration in ML workflow management.

For future work, we plan to evaluate the system's performance through user studies across stakeholders with varying technical backgrounds and domains, collecting metrics on task completion efficiency and usability. We will also focus on addressing edge cases identified during these evaluations, particularly improving the system's reasoning for complex multi-step operations and enhancing its ability to recover from ambiguous user instructions. Additionally, we intend to integrate a \emph{Code Agent} that will create new pipeline components and workflows from human instructions, dynamically compile them, and upload them to Kubeflow. This capability will transform the system from primarily supporting discovery and execution of existing pipelines to facilitating the creation of new workflows without requiring manual coding, further assisting ML development across research teams and disciplines.

\section*{Acknowledgment}
Part of the research leading to the results presented in this paper has received funding from the European Union’s funded Project HUMAINE under grant agreement no 101120218.

\bibliography{references}
\bibliographystyle{IEEEtran}

\end{document}